%% file: main.tex
\documentclass[sigconf,natbib=true,anonymous=false]{acmart}
\usepackage{algorithm}
\usepackage{algorithmic}
\def\tcr{\textcolor{red}}
\def\tcb{\textcolor{blue}}

\usepackage{multirow}
\usepackage[preprint]{}

\AtBeginDocument{%
  \providecommand\BibTeX{{%
    \normalfont B\kern-0.5em{\scshape i\kern-0.25em b}\kern-0.8em\TeX}}}

\setcopyright{acmlicensed}
\copyrightyear{2024}
\acmYear{2024}
\acmDOI{XXXXXXX.XXXXXXX}

\acmConference[SIGIR '24]{Make sure to enter the correct
  conference title from your rights confirmation emai}{July 14--18,
  2024}{Washington, D.C.}
\acmISBN{978-1-4503-XXXX-X/18/06}

\begin{document}

%%
%% The "title" command has an optional parameter,
%% allowing the author to define a "short title" to be used in page headers.
\title[CR-LT-KGQA: A KGQA Dataset Requiring Commonsense Reasoning and Long-Tail Knowledge]{CR-LT-KGQA: A Knowledge Graph Question Answering Dataset Requiring Commonsense Reasoning and Long-Tail Knowledge}

%%
%% The "author" command and its associated commands are used to define
%% the authors and their affiliations.
%% Of note is the shared affiliation of the first two authors, and the
%% "authornote" and "authornotemark" commands
%% used to denote shared contribution to the research.
\author{Willis Guo}
\authornote{Equal contribution}
\email{gwillis.guo@mail.utoronto.ca}
\affiliation{%
  \institution{University of Toronto}
  \city{Toronto}
  \state{ON}
  \country{Canada}
}

\author{Armin Toroghi}
\authornotemark[1]
\email{armin.toroghi@mail.utoronto.ca}
\affiliation{%
  \institution{University of Toronto}
  \city{Toronto}
  \state{ON}
  \country{Canada}}

\author{Scott Sanner}
\authornote{Affiliate to Vector Institute of Artificial Intelligence, Toronto}
\email{ssanner@mie.utoronto.ca}
\affiliation{%
  \institution{University of Toronto}
  \city{Toronto}
  \state{ON}
  \country{Canada}}

%%
%% By default, the full list of authors will be used in the page
%% headers. Often, this list is too long, and will overlap
%% other information printed in the page headers. This command allows
%% the author to define a more concise list
%% of authors' names for this purpose.
\renewcommand{\shortauthors}{Guo and Toroghi, et al.}

%%
%% The abstract is a short summary of the work to be presented in the
%% article.
\begin{abstract}
    \input{CONTENT/abstract} 
\end{abstract}

%%
%% The code below is generated by the tool at http://dl.acm.org/ccs.cfm.
%% Please copy and paste the code instead of the example below.
%%
\begin{CCSXML}
<ccs2012>
   <concept>
       <concept_id>10002951.10003317.10003338.10003341</concept_id>
       <concept_desc>Information systems~Language models</concept_desc>
       <concept_significance>500</concept_significance>
       </concept>
   <concept>
       <concept_id>10002951.10003317.10003347.10003348</concept_id>
       <concept_desc>Information systems~Question answering</concept_desc>
       <concept_significance>500</concept_significance>
       </concept>
   <concept>
       <concept_id>10010147.10010178.10010187</concept_id>
       <concept_desc>Computing methodologies~Knowledge representation and reasoning</concept_desc>
       <concept_significance>500</concept_significance>
       </concept>
 </ccs2012>
\end{CCSXML}

\ccsdesc[500]{Information systems~Language models}
\ccsdesc[500]{Information systems~Question answering}
\ccsdesc[500]{Computing methodologies~Knowledge representation and reasoning}

%%
%% Keywords. The author(s) should pick words that accurately describe
%% the work being presented. Separate the keywords with commas.
\keywords{Question Answering, Knowledge Graphs, Commonsense Reasoning, Large Language Models}

%% A "teaser" image appears between the author and affiliation
%% information and the body of the document, and typically spans the
%% page.

\received{20 February 2007}
\received[revised]{12 March 2009}
\received[accepted]{5 June 2009}

%%
%% This command processes the author and affiliation and title
%% information and builds the first part of the formatted document.
\maketitle

\section{Introduction}
\input{CONTENT/introduction}

\section{Related Work}
\input{CONTENT/related_work}

\section{Dataset}
\input{CONTENT/dataset}

\section{Experimental Setup}
\input{CONTENT/experimental_setup}

\section{Results}
\input{CONTENT/results}

\section{Conclusion}
\input{CONTENT/conclusion}

%%
%% The next two lines define the bibliography style to be used, and
%% the bibliography file.
\bibliographystyle{ACM-Reference-Format}
\bibliography{refs}

%%
%% If your work has an appendix, this is the place to put it.
\appendix

\end{document}

%% file: CONTENT/abstract.tex
Knowledge graph question answering (KGQA) is a well-established field that seeks to provide factual answers to natural language (NL) questions by leveraging knowledge graphs (KGs). However, existing KGQA datasets suffer from two significant limitations: (1) no existing KGQA dataset requires commonsense reasoning to arrive at an answer and (2) existing KGQA datasets focus on popular entities for which large language models (LLMs) can directly answer without hallucinating and without leveraging the KG. In this work, we seek a novel KGQA dataset that supports commonsense reasoning and focuses on long-tail entities (e.g., non-mainstream and recent entities) where LLMs frequently hallucinate, and thus create the need for novel methodologies that leverage the KG for factual and attributable commonsense inference. We create a novel Commonsense Reasoning (CR) and Long-Tail (LT) KGQA dataset with two subtasks---question answering and claim verification---that address both limitations (1) and (2). We construct CR-LT-KGQA\footnotemark[1] by building extensions to existing reasoning datasets StrategyQA and CREAK over Wikidata. While existing KGQA methods are not applicable due to their lack of commonsense inference support, baseline evaluation of LLMs on CR-LT KGQA demonstrate a high rate of hallucination. Thus, CR-LT KGQA poses significant challenges for hallucination-prone LLMs, hence paving the way for future commonsense KGQA research to provide accurate and factual answers for long-tail entities in the era of LLMs. 

%% file: CONTENT/introduction.tex
Knowledge graphs (KGs) provide explicit, structured representations of knowledge and store vast amounts of information about the world. However, querying KGs require expertise in complex formal query languages like SPARQL~\cite{kgqa-survey}, making them inaccessible. Knowledge graph question answering (KGQA)~\cite{zheng2017natural, berant2013semantic, yih2016value} is an important line of research whose ultimate goal is to democratize access to KGs by providing factual answers to natural language (NL) questions using KGs. While existing KGQA datasets have enabled substantial progress in KGQA to date, we observe two major limitations with existing KGQA datasets, that if addressed, should significantly expand the range of applicability of KGQA methods.
%we believe will make KGQA even more applicable and useful than it already is. 

First, no existing KGQA dataset requires commonsense reasoning to arrive at an answer. Instead, questions are purely factoid, for instance, \textit{“Who wrote the 1956 novel 101 Dalmatians?”} Such factoid questions are the most obvious types of questions that can benefit from being grounded in KGs. However, there is a rich space of questions that go beyond facts and require commonsense reasoning to answer. For instance, the question \textit{“Did any of Ibrahim Moustafa’s children follow in his footsteps?”} requires commonsense reasoning to infer that \textit{to follow in one’s footsteps} means to pursue a similar career. Having made this commonsense inference, KGs then provide the necessary facts to answer the question. While such implicit reasoning is trivial to humans, endowing AI systems with human-like commonsense reasoning capabilities remains an elusive goal. 

Second, existing KGQA datasets focus on popular entities that LLMs can answer using their internal knowledge without hallucinating and without needing a KG. That is, LLMs have fundamentally changed QA, often obviating the need for a KG. However, LLMs are prone to hallucinate when long-tail knowledge with little support in the LLM’s pre-training data is required. It is precisely in these long-tail settings where KGs are vital for grounding LLMs. Currently, no KGQA dataset evaluates long-tail knowledge settings, the intersection of where LLMs fail and KGs are required. 

Thus, both commonsense reasoning and long-tail knowledge are key characteristics of real-world questions that existing KGQA datasets do not support. In this work, we introduce CR-LT-KGQA,\footnote{\href{https://github.com/D3Mlab/cr-lt-kgqa}{https://github.com/D3Mlab/cr-lt-kgqa}} a novel KGQA dataset with two subtasks, question answering and claim verification, that addresses both aforementioned limitations of existing KGQA datasets. CR-LT-KGQA requires (1) commonsense reasoning that goes beyond the factoid questions in existing KGQA datasets, and (2) long-tail entity knowledge which exacerbates LLM hallucinations and thus requires strict reasoning over the KG. 

We construct CR-LT-KGQA by building extensions to two existing reasoning datasets, StrategyQA~\cite{strategyqa} and CREAK~\cite{creak}, over Wikidata{\footnote{\url{https://www.wikidata.org/}}}. Neither of these is a KGQA datasets; instead, they are NL commonsense reasoning datasets with queries about mostly famous persons, places, events, etc. In this paper, we utilize the commonsense reasoning element of these datasets and transform their queries to form CR-LT-KGQA, the first KGQA dataset requiring commonsense reasoning and targeting long-tail knowledge.

Baseline evaluations on CR-LT-KGQA demonstrate that LLMs suffer from a high rate of hallucination in long-tail settings. Thus, CR-LT-KGQA represents a novel evaluation benchmark that considers both the opportunities and challenges presented by LLMs, paving the way for future KGQA research in the new era of LLMs. In summary, our key contributions are as follows: 
\begin{itemize}
    \item We create CR-LT-KGQA, a novel dataset requiring commonsense reasoning and long-tail knowledge about entities. 
    \item We establish baseline results, which leave much room for improvement, thus inciting future KGQA research to design methodologies capable of handling the complex queries in CR-LT-KGQA. 
\end{itemize}

%% file: CONTENT/related_work.tex
\subsection{KGQA Datasets}
% Given infinite time, examples of questions for each dataset (perhaps in a master table would be interesting).

Early KGQA datasets consisted of simple questions that can be answered using a single KG triple. Recent efforts in KGQA dataset construction have focused on generating more complex questions that involve, for instance, multi-hop reasoning. 

\vskip 4pt 
\noindent
\textbf{WebQuestions}~\cite{webquestions} is grounded in Freebase~\cite{freebase}, and is built by obtaining single-entity questions using the Google Suggest API and using crowd workers to pick the ones that are answerable using only the Freebase page for the question's subject entity. WebQuestionsSP~\cite{webquestions-sp} is an extension of WebQuestions with SPARQL annotations.

\vskip 4pt 
\noindent
\textbf{LC-QuAD}~\cite{lcquad} is grounded in DBpedia~\cite{lehmann2015dbpedia}, and is built by: (1) generating formal queries using query templates as well as selected entities and predicates, (2) rewriting them in NL using NL templates written for each query template, and (3) asking crowdworkers to rewrite the NL template into a NL question.

\vskip 4pt 
\noindent
\textbf{GrailQA}~\cite{grailqa} is grounded in Freebase, and is built using a 4-step process: (1) algorithmically generate logical forms from a KG, (2) convert them to pseudo-NL questions, (3) use crowdsourcing to paraphrase the pseudo-NL questions into NL questions and (4) use crowdsourcing to cross-validate the NL questions. 

\vskip 4pt 
\noindent
The most common approach for building KGQA datasets is to first write logical forms and then rewrite them as NL questions. Since logical forms are meant to be executed over KGs, they are by nature factoid, and thus the resulting questions will naturally be factoid as well, no matter how complex the logical form. However, there is a rich space of questions requiring commonsense reasoning that is not yet covered in any existing KGQA dataset.

\subsection{QA Datasets}
% Given infinite time, examples of questions for each dataset (perhaps in a master table would be interesting).

Some non-KG QA datasets do require more complex forms of reasoning including commonsense reasoning. However, they focus on popular entities that LLMs can recall with little hallucination. 

\vskip 4pt
\noindent
\textbf{CommonsenseQA}~\cite{commonsenseqa} is a multiple-choice commonsense QA dataset. Crowd workers are given a source concept and target concepts related to the source concept extracted from ConceptNet~\cite{concept-net}, a commonsense knowledge graph, and asked to write questions that discriminate between the target entities. 

\vskip 4pt
\noindent
\textbf{StrategyQA}~\cite{strategyqa} is a true/false QA dataset where the required reasoning steps are implicit and should be inferred using a strategy. Each query is annotated with the reasoning steps and the evidence Wikipedia paragraphs to answer each step. Crowd workers are asked to write strategy questions given an entity term, and adversarial models are continuously trained to collect increasingly difficult questions.

\vskip 4pt
\noindent
\textbf{CREAK}~\cite{creak} is a true/false QA dataset that requires commonsense reasoning about entities to answer the queries. Similar to StrategyQA, crowd workers are asked to write queries given an entity term. However, unlike StrategyQA, a model-in-the-loop approach is not used. 

\vskip 4pt 
\noindent
% The following is a strong claim to make without citations unless you state "The most common approach for building **the above** QA datasets"
The common underlying approach for building the above QA datasets requiring is to let crowd workers write questions given one or more entities. Although this approach of priming crowd workers gives them more freedom to write diverse and challenging questions that require implicit reasoning, the chosen starting entities are often well-known entities.

%% file: CONTENT/dataset.tex
\begin{figure*}[!t]
  \centering
  \includegraphics[width=0.9\linewidth]{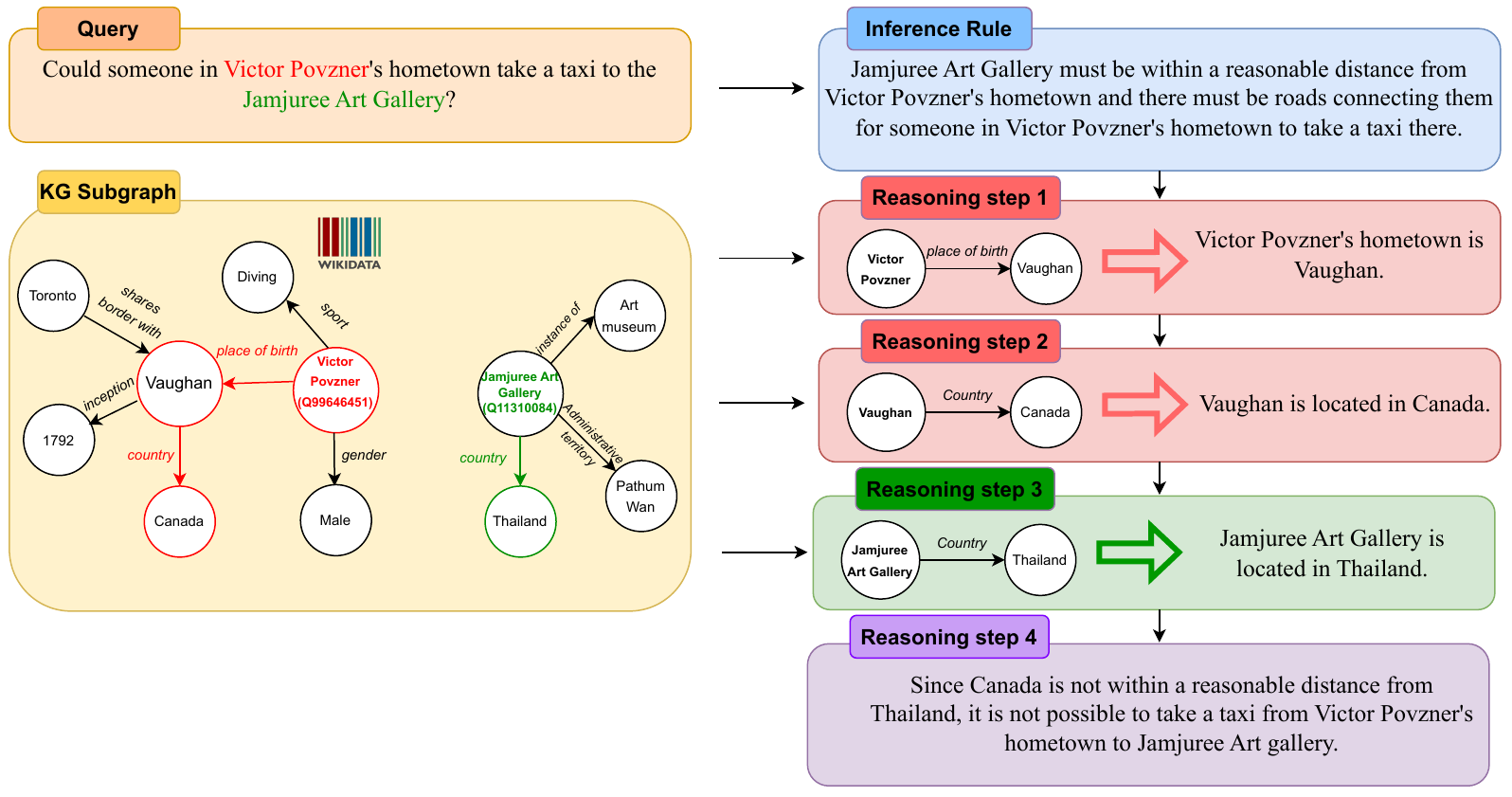}
    \caption{A simple example of an entry from CR-LT-KGQA. Each entry consists of (i) query (ii) Wikidata QIDs of anchor entities, (iii) relevant Wikidata sub-graph, (iv) inference rule, and (v) reasoning steps.}
      \label{crlt}
\end{figure*}

\subsection{Description}
%To address the shortcomings of existing KGQA datasets, we propse CR-LT-KGQA, the first KGQA dataset that requires commonsense reasoning and focuses on long-tail entities.
CR-LT-KGQA has a total of 350 queries and consists of two subsets for question answering, and claim verification tasks. The question answering subset includes 200 questions and the claim verification subset contains 150 claims formed around entities of Wikidata KG belonging to long-tail knowledge. An exemplar entry from CR-LT-KGQA is shown in Figure \ref{crlt}. Each entry of the dataset  consists of (i) a query in the form of a question or a claim (ii) the unique Wikidata QIDs of anchor entities mentioned in the query, (iii) the relevant Wikidata sub-graph, (iv) an inference rule expressing the commonsense knowledge required for answering the query in the form of an NL axiom, and (v) reasoning steps grounded on KG triples to conclude the final answer. In this section, we explain each constituent of CR-LT-KGQA entries in more detail.

\vskip 4pt
\noindent 
\textbf{Query}
A query in CR-LT-KGQA is either in the form of a question expressed as an interrogative sentence, or a claim expressed as a declarative sentence. We use the term \textit{query} to refer to both questions and claims. Each query $q$ has a ground truth answer $a_q \in \{True, False\}$ which can be found by leveraging the facts from the KG and commonsense reasoning.

\vskip 4pt
\noindent 
\textbf{Wikidata entities.} As in other KGQA datasets, each query in CR-LT-KGQA is formed around a number of KG entities. The query mentions a set of anchor entities whose KG triples are necessary to begin the reasoning process. For each anchor entity, we provide both the label (name) and the QID which uniquely identifies the entity in Wikidata. These QIDs and labels are used to retrieve the relevant facts required for answering the queries. 
\vskip 4pt
\noindent 
\textbf{Inference rule.}
The inference rule is a logical statement in the form of an NL axiom that formally expresses the commonsense knowledge that is required to answer the question or verify the claim. In contrast to the existing datasets that consider inference using commonsense knowledge as an implicit part of the reasoning process, in order to enhance verifiability, we explicitly annotate the entry with the inference rules that are required to identify the answer to each question or claim. These inference rules form the set of properties of entities targeted in the query and relations among them as conditions and premises that will lead the reasoning process, such that if all premises are true, then the answer is true but if any premise is false, then the answer is false.

Formally, Denoting the set of entities targeted in a question or claim $q$ by $\mathcal{E}_{q} = \{e_{1,q}, ..., e_{|\mathcal{E}_{q}|,q}\}$, inference rule $I_q$ is an NL representation of the First-Order Logic (FOL) expression
\begin{equation}
\left( \bigwedge\limits_{i=1}^{|\mathcal{P}|}\bigwedge\limits_{j=1}^{|\mathcal{E}|}P_i(e_j) \right) \land \left( \bigwedge\limits_{i=1}^{|\mathcal{F}|}\bigwedge\limits_{j=1}^{|\mathcal{E}|}F_i(e_j) 
\, \langle\mathit{op}_j^i\rangle \, e_j^i \right) \implies a_q,
\end{equation}
in which $\mathcal{P} = \{ P_1, ..., P_{|P|} \}$ is the set of predicates, $\mathcal{F} = \{ F_1, ..., F_{|F|} \}$ is the set of functions, $\langle\mathit{op}_j^i\rangle \in \{ =, \neq, <, \leq, >, \geq \}$ is a (dis)equality operator or comparison operator if the function value is numeric, $e_j^i$ is the entity compared to the function evaluation, and $a_q$ is the answer to the query or claim. 

 The assignment of each property and function to an entity is performed using one of the formats of \textit{"\{entity\} must \{have property or properties\} to \{conclusion\}"} or \textit{"If \{entity has the property\}, then \{the conclusion\}"} whichever results in a sentence that sounds more natural. For example, \textit{"If Rodney Strasser is a girl from Latin America and is about to celebrate her 15th birthday, it would make sense for her to ask for a quinceañera."}

%of the form \textit{<TODO: inference rule form>}. The inference rule is critical for answering the question because it expresses the commonsense inference required to arrive at a factual and verifiable premise that is different than the question such that if the premise is true, then the answer is true but if the premise is false, then the answer is false. 

\vskip 4pt
\noindent 
\textbf{Relevant KG Sub-graph.}
Queries in CR-LT-KGQA are curated by ensuring that all factual information required for answering them are present in the Wikidata KG. For each query, we provide the set of all triples in the relevant KG sub-graph of Wikidata that supply the required factual knowledge. Wikidata triples can also include qualifiers in the format: \textit{relation-entity} pairs that provide additional context to the original KG triple. We represent these hyper-relational triples as $\big((h, r, t), \{(k_i, v_i)\}_{i=1}^m\big)$ in which $h$, $r$, and $t$ denote the head, relation, and tail of the original triple, and $k_i$ and $v_i$ denote the relation and entity of the quantifier respectively. For example, for KG triple \textit{(Virginia Raggi, position held, mayor of Rome)}, an additional quantifier \textit{(replaces, Ignazio Marino)} is provided in Wikidata. We present the triple together with the quantifier as \textit{(Virginia Raggi, position held, mayor of Rome \{replaces, Ignazio Marino\})}.
\vskip 4pt
\noindent 
\textbf{Reasoning steps.}
We decompose the commonsense inference rule into an ordered sequence of $n$ intermediate reasoning steps $S = (s^{(1)}, \dots, s^{(n)})$ for verifying whether or not each premise of the commonsense inference rule holds. There are two types of reasoning steps: (i) Steps involving identifying the relevant facts among the facts retrieved from Wikidata, and (ii) Steps in which logical reasoning about the extracted facts is performed. Each reasoning step $s^{(i)}$ is written as a declarative sentence and may depend on the result of earlier steps $s^{(j)}, j < i$. Also, for each reasoning step $S^{(i)}$ that involves extraction of relevant Wikidata facts, we provide the ground truth evidence Wikidata triples $T^{(i)} = (T^{(i)}_1, \dots, T^{(i)}_m)$ that will help in evaluating the factuality of answers at a step-by-step level.
%We decompose the commonsense inference rule into an ordered sequence of $n$ intermediate reasoning steps $S = (s^{(1)}, \dots, s^{(n)})$ for verifying whether or not the premise of the commonsense inference rule is true. There are two types of reasoning steps: retrieving required facts from Wikidata, and logical reasoning about the required facts. Each reasoning step $s^{(i)}$ is written as a question, and may depend on the result of earlier steps $s^{(j)}, j < i$.

%For each reasoning step $S^{(i)}$ that involves retrieving facts from Wikidata, we provide the ground truth evidence Wikidata triples $T^{(i)} = (T^{(i)}_1, \dots, T^{(i)}_m)$. Wikidata triples can contain qualifiers: key-value relation-entity pairs that provide additional context to the original KG triple. We represent these hyper-relational triples as: $\big((h, r, t), \{(k_i: v_i)\}_{i=1}^m\big)$

\vskip 4pt
\noindent
Unlike some existing KGQA datasets \cite{grailqa, webquestions-sp, lcquad}, we do not annotate CR-LT KGQA with ground truth formal queries. This is because our questions require making commonsense inferences and multiple steps of reasoning, and thus there is no single formal query that can directly answer the question.

\begin{table*}[t!]
\caption{Exemplar queries from the StrategyQA and Creak datasets and their counterparts in CR-LT-KGQA: \tcr{red parts} are logic or clarity problems due to invalid assumption or unnatural wording, and \tcb{blue parts} are lack of entity or not being long-tail knowledge}.
\small
\centering
\begin{tabular}{|l|l|}
\hline
\multicolumn{2}{|c|}{\textbf{Question Answering (Example 1)}} \\ 
\hline
 \textbf{Original Query} & \tcr{Did} \tcb{Francois Mitterrand} ever meet \tcb{Barak Obama} while they both held the position of President?\\ 
 \hline
 \textbf{CR-LT Query}&  \tcr{Could} \tcb{Liau Hiok-hian} and \tcb{Virginia Raggi} have met while they both held the position of council member?\\
 \hline
 \textbf{Modifications} & \tcr{1-} The original question intends to ask the possibility of occurrence of a meeting rather than whether the meeting actually took\\ & place, but phrases it in a misleading manner.\\
 &\tcb{2-} The famous entities are replaced with long-tail counterparts.\\
 \hline
\multicolumn{2}{c}{}\\
%\end{tabular}
% \end{table*}

% \begin{table*}[t!]
% \small
% \centering
%\begin{tabular}{|c|c|}
\hline
\multicolumn{2}{|c|}{\textbf{Question Answering (Example 2)}} \\ 
\hline
 \textbf{Original Query} & \tcr{Could} \tcb{Tom Cruise} \tcr{explain} mental auditing?\\ 
 \hline
 \textbf{CR-LT Query}&  \tcr{Is it likely} for \tcb{Julia Nickson-Soul} to \tcr{be familiar with} mental auditing ?\\
 \hline
 \textbf{Modifications} & \tcr{1-} The original question makes the implicit assumption that every person who practices a religion is able to explain its terminology \\&  which is not necessarily True. In the modified query, we replace it with \textit{being likely to be familiar} which is more accurate.\\
 &\tcb{2-} The famous entities are replaced with long-tail counterparts.\\
 \hline
\multicolumn{2}{c}{}\\
% \end{tabular}
% \end{table*}

% \begin{table*}[t!]
% \small
% \centering
% \begin{tabular}{|c|c|}
\hline
\multicolumn{2}{|c|}{\textbf{Claim Verification (Example 1)}} \\ 
\hline
 \textbf{Original Query} & \tcb{All nuns} \tcr{act in holy ways}.\\ 
 \hline
 \textbf{CR-LT Query}& \tcb{ Léocadie Gascoin} \tcr{considered her job} to be holy.\\
 \hline
 \textbf{Modifications} & \tcr{1-} The original question is vague. 
 \textit{Acting in holy ways} is not a clear expression. In the modified query, we replaced it with a \\ & clear question.\\
 &\tcb{2-} The question does not target a particular entity. In the modification, we targeted it on a long-tail entity.\\
 \hline
\multicolumn{2}{c}{}\\
% \end{tabular}
% \end{table*}

% \begin{table*}[t!]
% \small
% \centering
% \begin{tabular}{|c|c|}
\hline
\multicolumn{2}{|c|}{\textbf{Claim Verification (Example 2)}} \\ 
\hline
 \textbf{Original Query} & \tcb{The civil engineer} designed \tcr{the new suspension bridge} in \tcr{the city}.\\ 
 \hline
 \textbf{CR-LT Query}&  \tcb{Władysław Folkierski} had \tcr{likely learned} the knowledge required to design \tcr{suspension bridges}.\\
 \hline
  \textbf{Modifications} & \tcr{1-} The original claim is vague. It is not clear what civil engineer it refers to and targets a specific event for which no evidence is \\ &provided.\\
 &\tcb{2-} The original claim does not anchor a specific KG entity, but in the modification we fixed this.\\
 \hline
\end{tabular}
\label{modifications}
\end{table*}

\subsection{Method}\label{method}

\vskip 4pt
\noindent
\textbf{Query selection.} To generate CR-LT-KGQA queries, we first select questions from StrategyQA and claims from CREAK for which the required factual knowledge for answering them is present in Wikidata or that can be rewritten as such queries by targeting them on new KG entities. We follow a two-step process with two annotators. The first annotator selects the query, and the second annotator validates the first annotator’s question selection by independently verifying that Wikidata indeed contains the facts required to answer the question. For StrategyQA, since questions are annotated with strategy decompositions, we maintained priority for questions with greater numbers of strategy decompositions due to their more challenging reasoning procedure. For CREAK, claims with longer explanations involving more diverse reasoning skills were given higher priority over the simple ones. In general, the annotators observed that claims of Creak generally required an easier reasoning process compared to those of StrategyQA. 

\vskip 4pt
\noindent
\textbf{Entity substitution.} CR-LT-KGQA queries are designed to focus on obscure entities from long-tail knowledge. To achieve this, we replace the original well-known entities with long-tail counterparts of the same type as the original entity. Specifically, we found similar counterparts of existing entities by manually generating SPARQL queries on Wikidata to retrieve candidate entities with similar properties to the entities in the original query. Next, among the candidate entities, an entity with a considerably smaller number of Wikidata triples was randomly selected. Finally, this process was often followed by a Google search to ensure that the new entity is less famous than the original one by viewing the amount of search results. We also modify cases in which the query is not targeted at a KG entity by introducing a long-tail KG entity to it.  

Examples of the entity replacement are provided in the blue-colored parts of Table \ref{modifications}. As indicated in these examples, famous entities like \textit{Francois Mitterrand} and \textit{Barak Obama} are replaced with less well-known entities while preserving their type which in these cases are all \textit{humans} and \textit{politicians}. Also, for queries with no particular entity such as \textit{all nuns} or \textit{The civil engineer}, we introduce long-tail entities of that type and target the query on them, which in these cases are \textit{a nun} and \textit{a civil engineer}.

\vskip 4pt
\noindent
\textbf{Question rewriting.} \citet{why-would-you-ask-it-that-way} found that questions in existing KGQA datasets are often worded unnaturally, and that consequently the performance of KGQA methods drop when evaluated on rewritten, more natural formulations of the same questions. We follow their coding scheme for evaluating the naturalness of NL Questions, which includes dimensions like grammar (e.g., poor word ordering, non-idiomatic) and form (e.g., quizlike, imperative phrasing), to rewrite more natural formulations of our questions. We also observe that some queries in Creak and StrategyQA are written with making implicit assumptions that are not necessarily correct. For instance, in the second example for Question answering in Table \ref{modifications}, an incorrect assumption is made. We try to remove these errors and flaws in the modified queries too. Examples of these issues and the modifications made are provided in the red-colored Table \ref{modifications}.

\subsection{Characteristics}
\textbf{Entity popularity.} To characterize the long-tail nature of our dataset, we plot the distribution of entity popularity for CR-LT versus StrategyQA and CREAK. Previous works measure the number of occurrences of an entity in a pretraining corpus~\cite{lm-long-tail, lm-memorization, pretraining-term-freq}, which is computationally expensive, or the page view count for the Wikipedia page of that entity~\cite{popqa}. Since our dataset is grounded in Wikidata, as explained in section \ref{method}, we use the number of Wikidata triples containing the entity as a heuristic measure for entity popularity. 

\begin{figure}
  \centering
  \includegraphics[width=0.9\linewidth]{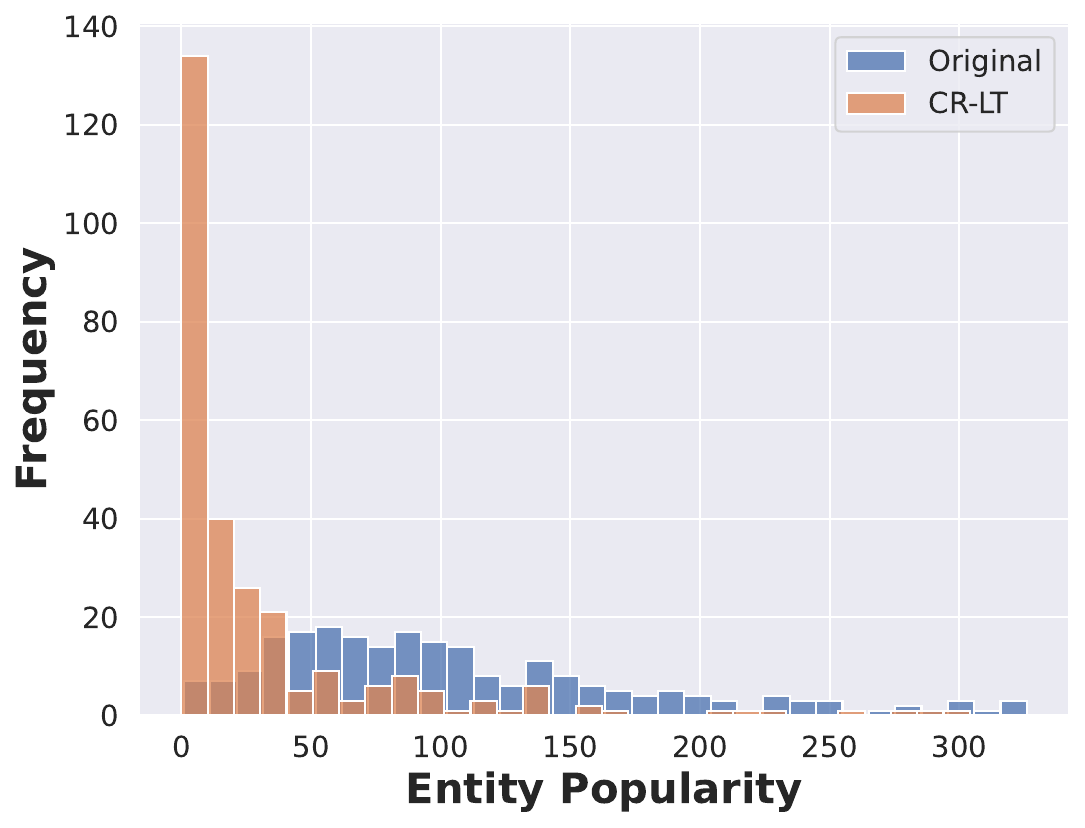}
  \caption{Distribution of entity popularities for the question answering subtask.}
  \label{strategyqa-entity-popularity}
\end{figure}

\begin{figure}
  \centering
  \includegraphics[width=0.9\linewidth]{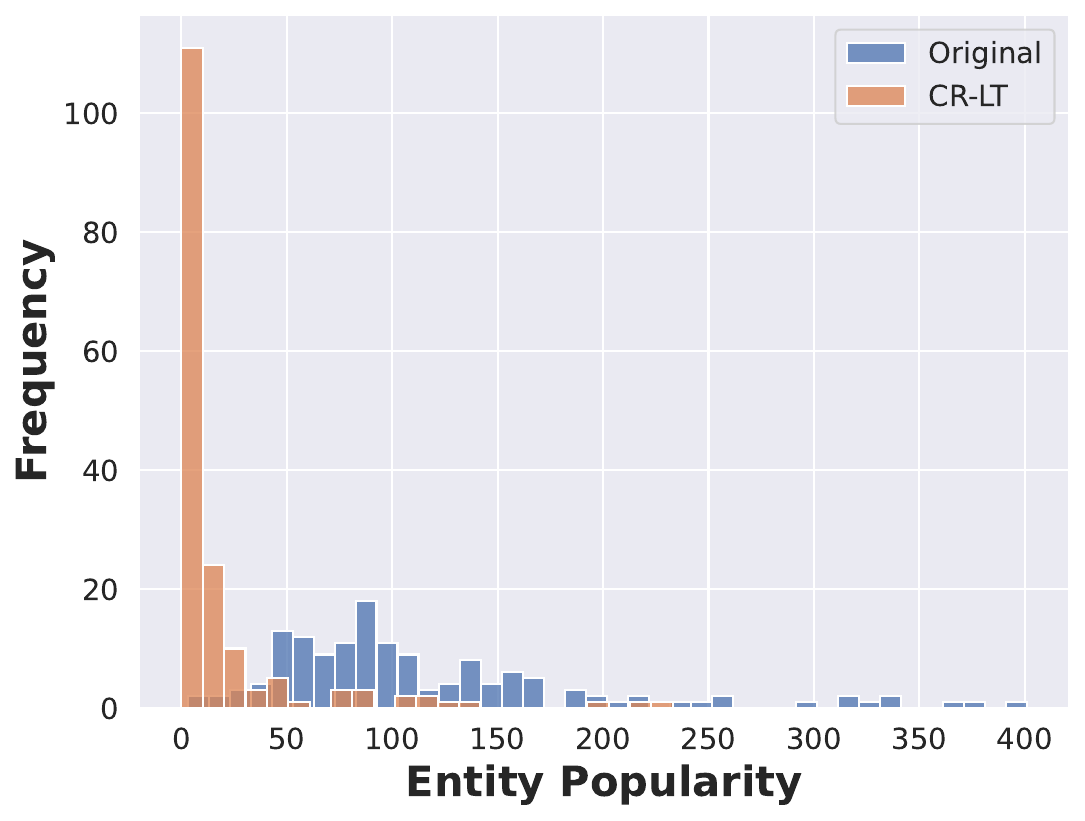}
  \caption{Distribution of entity popularities for the claim verification subtask.}
  \label{creak-entity-popularity}
\end{figure}

The distributions of entity popularity in the question answering and claim verification subtasks are provided in Figure \ref{strategyqa-entity-popularity} and Figure \ref{creak-entity-popularity} respectively. Evidently, the popularity of CR-LT-KGQA entities in both tasks of are considerably less than the entities of original queries which verifies its focus on long-tail knowledge. 

\begin{figure}
  \centering
  \includegraphics[width=0.9\linewidth]{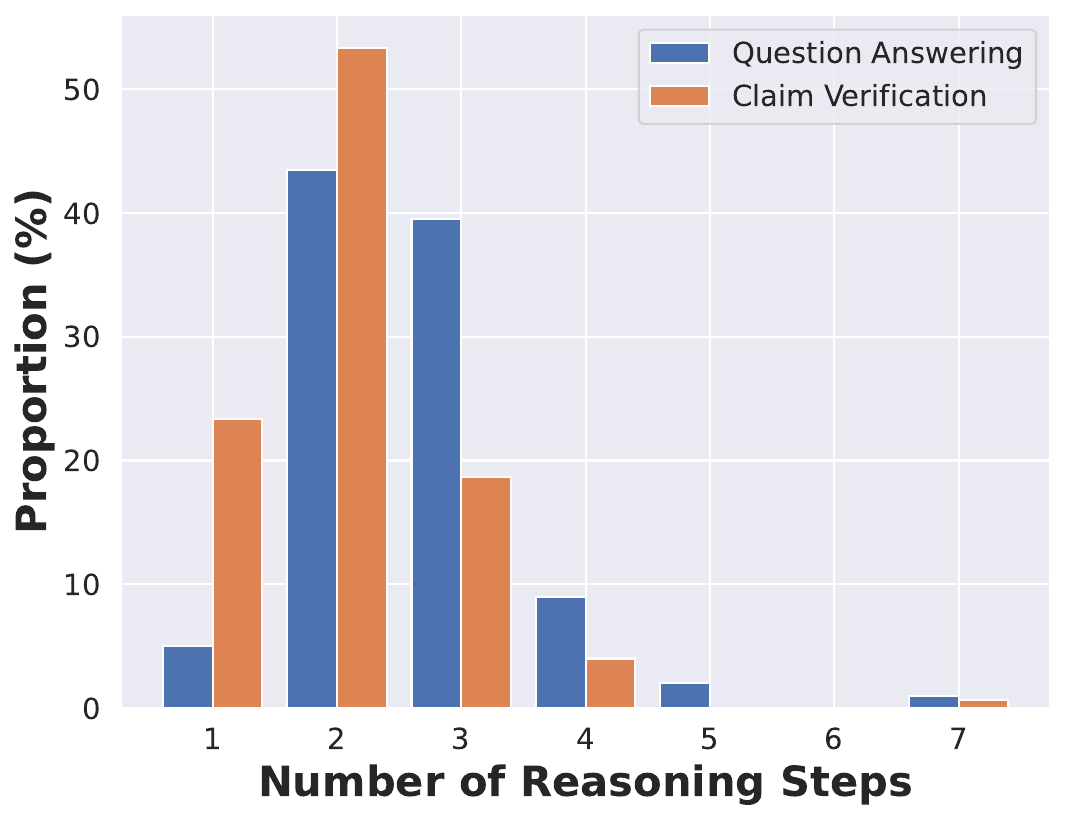}
  \caption{Distribution of the number of reasoning steps in the dataset subtasks.}
  \label{reasoningsteps}
\end{figure}

\vskip 4pt
\noindent
\textbf{Reasoning Steps.}
The number of reasoning steps required for determining the answer of a query can be used as a measure of its level of difficulty. The distribution of the number of reasoning steps on both tasks are provided in Figure \ref{reasoningsteps}. In general, the number of steps for both tasks are centered around 2 or 3 steps, with the question-answering task being more challenging than claim verification.

\vskip 4pt
\noindent
\begin{figure}
  \centering
  \includegraphics[width=0.8\linewidth]{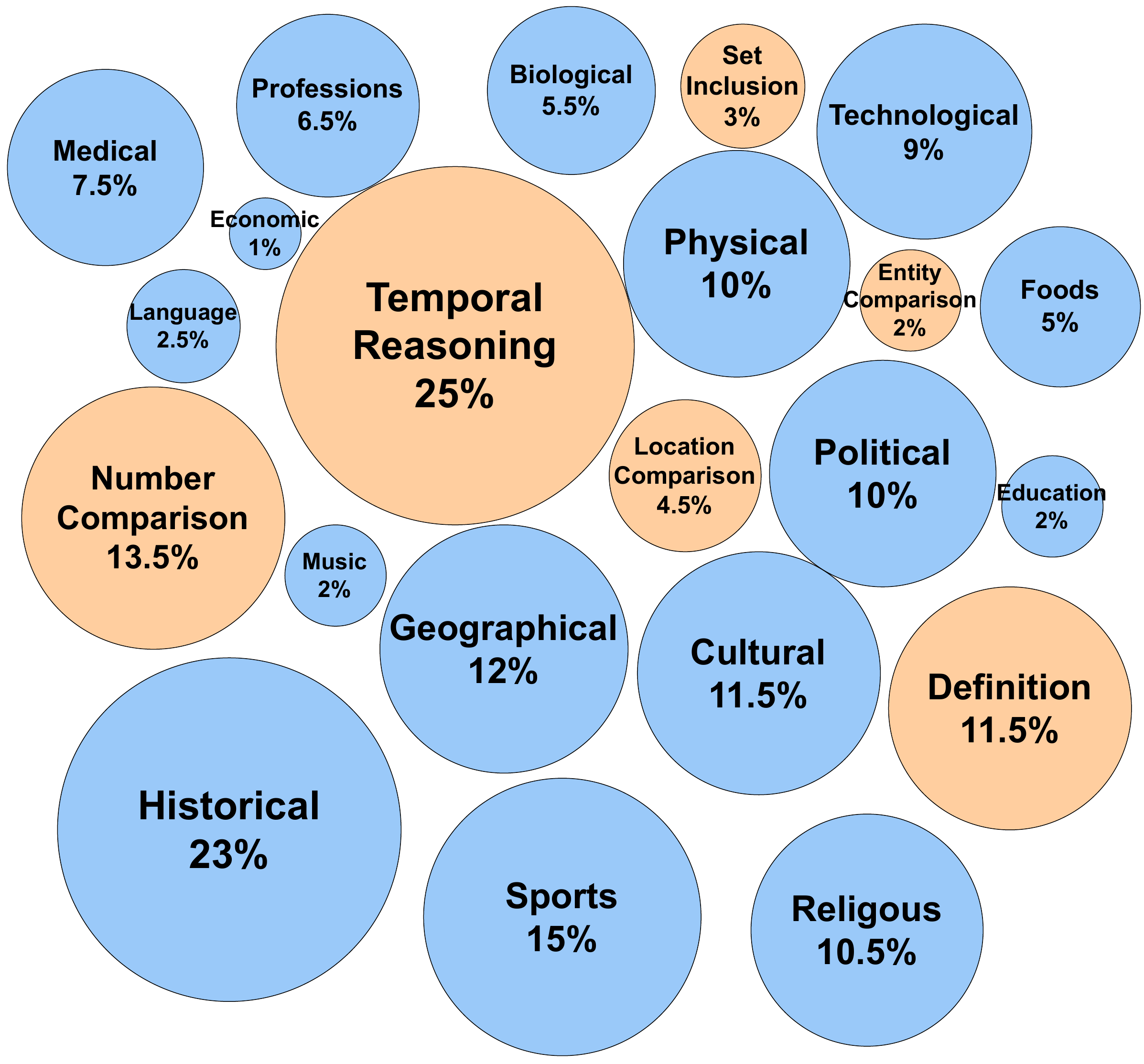}
    \caption{Distribution of reasoning skills in question answering task. Orange (Blue) circles show domain-independent (domain-dependent) reasoning skills.}
  \label{skillsqa}
\end{figure}
\vspace{-3mm}
\begin{figure}
  \centering
  \includegraphics[width=0.8\linewidth]{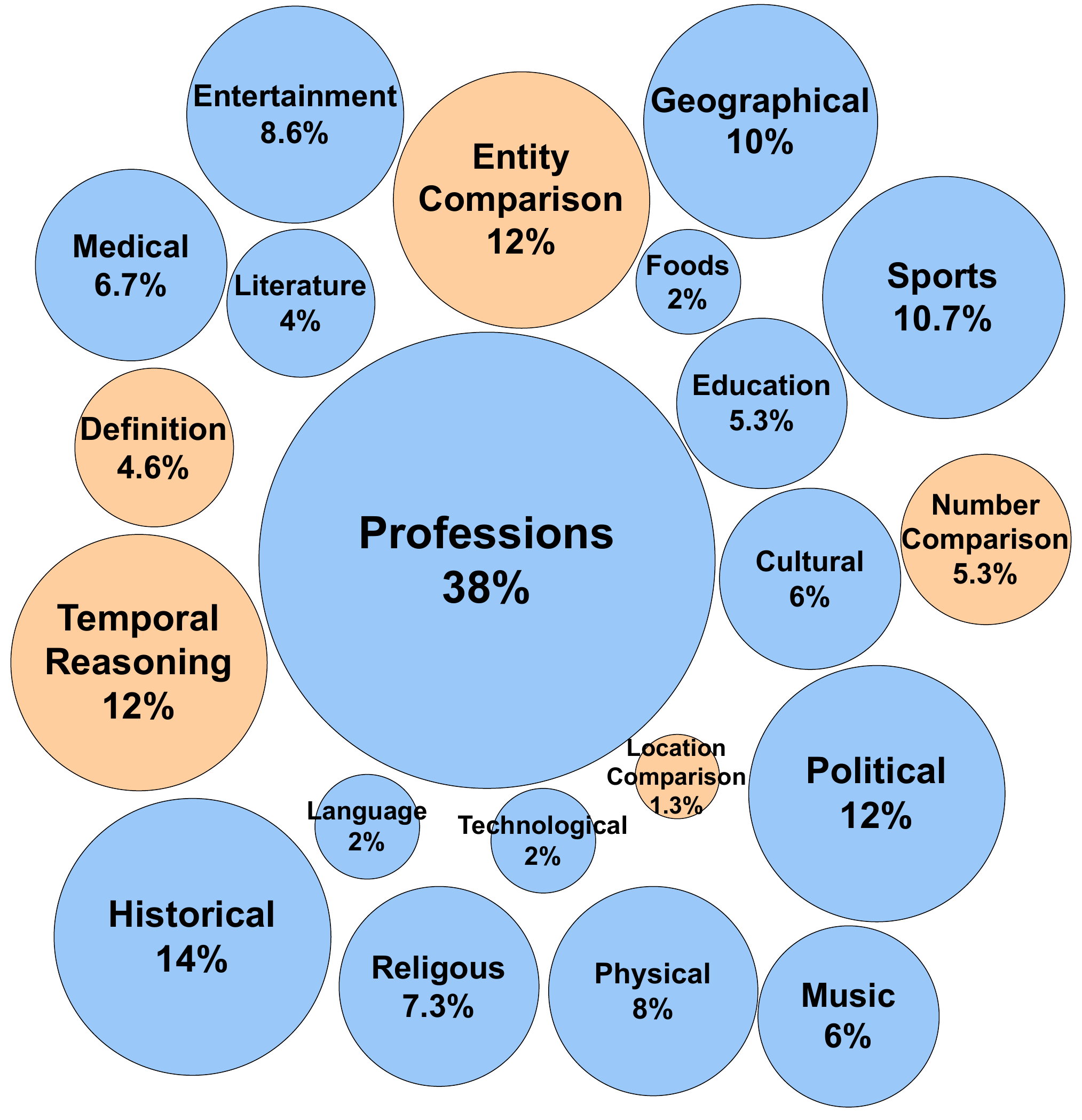}
\caption{Distribution of reasoning skills in claim verification task. Orange (Blue) circles show domain-independent (domain-dependent) reasoning skills.}
  \label{skillscreak}
\end{figure}

\vskip 4pt
\noindent
\textbf{Reasoning Skills} 
In order to explore the diversity of reasoning skills required for solving the queries, following~\citep{strategyqa}, we annotate the set of skills involved in the reasoning steps for each entry. These skills can be broadly categorized into domain-specific skills that require knowledge about a particular domain such as geography or politics, and domain-independent skills such as temporal reasoning or number comparison. The distribution of the required skills in question answering and claim verification task are provided in Figure \ref{skillsqa} and \ref{skillscreak} respectively. These figures show the broad coverage of the necessary reasoning skills across both tasks. Examples of queries for the five most frequent skills in question answering and claim verification tasks are provided in Table \ref{skillstableqa} and Table \ref{skillstablecreak} respectively.

\begin{table}[t!]
\caption{Most frequent reasoning skills required for question answering task and exemplar questions.}
\centering
\begin{tabular}{|c|c|} 
\hline
\textbf{Reasoning Skill} & \textbf{Exemplar Question}\\
\hline
Temporal Reasoning & \textit{Could Giovanni Battista Crespi have }\\ &   \textit{read L'apostolo popolare?} \\
\hline
Historical & \textit{Would it have been possible for }\\ &\textit{Maria de Ventadorn to speak to someone}\\  & \textit{100 miles away?}\\
\hline
Sports & \textit{Could Kim Dae-jung form}\\ &\textit{a polo team from his children?}\\
\hline
Number Comparison & \textit{Did Dariush Homayoun have longer}\\ & \textit{longevity than Bozorg Alavi?}\\
\hline
Geographical & \textit{Is Lowshan located south of Gubadly?}\\
\hline
\end{tabular}
\label{skillstableqa}
\end{table}

\begin{table}[t!]
\caption{Most frequent reasoning skills required for claim verification task and exemplar claims.}
\centering
\begin{tabular}{|c|c|} 
\hline
\textbf{Reasoning Skill} & \textbf{Exemplar Claims}\\
\hline
Professions & \textit{Martina Di Bari mostly uses her }\\ &   \textit{foot when doing her job.} \\
\hline
Historical & \textit{María Subíes Forcada's home}\\ &\textit{country was among the allied powers.}\\
\hline
Temporal Reasoning & \textit{Giovanni Battista Casti's works}\\ &\textit{may be influenced by}\\ &\textit{ Maria Grazia Lenisa's poems.}\\
\hline
Political & \textit{Hussein Ali Shido's party}\\ & \textit{supported anti-capitalism.}\\
\hline
Entity Comparison & \textit{Bogna Sobiech married an athlete who}\\ & \textit{does a different sport from her.}\\
\hline
\end{tabular}
\label{skillstablecreak}
\end{table}

%% file: CONTENT/experimental_setup.tex
We evaluate LLMs on both the original and CR-LT queries to evaluate the impact of long-tail knowledge for QA. We use GPT-3.5 Turbo with Chain-of-Thought (CoT) prompting, both in zero-shot~\cite{0-cot} and few-shot ($k=2$) settings ~\cite{cot}. We do not evaluate existing KGQA methods because they are limited to factoid queries and thus do not actually support queries requiring commonsense reasoning. For instance, neural semantic parsing methods like KB-BINDER~\cite{kb-binder} that generate logical forms will not work on CR-LT queries since there is no ground truth logical form that directly answers the query to begin with. Instead, our queries require making commonsense inferences, and then performing multiple steps of reasoning.  

% \subsection{Baselines}

% Our baselines include (1) pure LLM, to evaluate how well LLMs can accurately recall knowledge about long-tail entities, and (2) KGQA methods, to evaluate how well existing KGQA methods can answer questions that require commonsense reasoning. 

% \vskip 4pt
% \noindent 
% \textbf{LLM.} We use GPT-3.5 Turbo with CoT prompting, both in zero-shot and few-shot settings. 

% \vskip 4pt
% \noindent
% \textbf{KAPING.} A zero-shot, retrieval-augmented method. First, entities in the question are matched to KG entities using entity linking techniques. Second, relevant KG triples associated with those entities are added to the LLM prompt using dense retrieval. We use GPT-3.5 Turbo as the LLM.

\subsection{Metrics}
In addition to accuracy, we perform human evaluation using the following metrics to evaluate factual and reasoning faithfulness. 

\vskip 4pt
\noindent
\textbf{FActScore.} Factual precision in Atomicity Score~\cite{factscore} measures the percentage of atomic facts in the LLM's response that are supported by a given knowledge source $C$, in our case Wikidata. Given a LLM's response $y$ consisting of a set of atomic facts $A_y$, the FActScore is defined as:
\begin{equation}
    f(y) = \frac{1}{|A_y|} \sum_{a \in A_y} \mathbb{I}(a \text{ is supported by } C)
\end{equation}

\vskip 4pt
\noindent
\textbf{Reasoning score.} We define an analogous metric as FActScore, but for reasoning faithfulness. Given a LLM's response $y$ consisting of a sequence reasoning steps $S = (s_1, \dots, s_n)$, step $s_i$ is valid $V(s_i) = 1$ if and only if $s_i$ can be logically deduced from all previous steps $s_j, j < i$. Thus, $y$ is correct if and only if all intermediate steps are correct $V(S) = \wedge_{i=1}^n V(s_i)$

%% file: CONTENT/results.tex
\input{Tables/results}

Baseline results for both the question answering and claim verification subtasks of CR-LT are presented in table \ref{results}.

\subsection{Question Answering} 

\vskip 4pt
\noindent
\textbf{Overall.} 0-shot and 2-shot CoT achieve the same accuracy (0.7) on the original queries. Their answer rates are also similar: 0.89 for 0-shot and 0.92 for 2-shot CoT. However, in the long-tail CR-LT setting, both the accuracy and answer rate drop significantly. For instance, 0-shot CoT accuracy drops from 0.7 to 0.32, and the answer rate drops from 0.89 to 0.41. These results demonstrate that LLMs can adeptly answer queries about popular entities, but struggle to recall long-tail knowledge. Thus, CR-LT poses significant challenges for LLMs.

\vskip 4pt
\noindent
\textbf{Factuality.} LLMs achieve low FaCTScores, 0.54 and 0.52 for 0-shot and 2-shot CoT respectively, on CR-LT queries. As with overall accuracy and answer rate, this represents a significant drop in performance compared to the original queries. For instance, the FaCTScore for 0-shot CoT decreases from 0.63 to 0.54. These results crucially demonstrate that in the long-tail setting, even when LLMs do answer, there is a high rate of hallucination. Thus, CR-LT, unlike existing QA datasets, requires strict use of the KG and cannot be answered using the internal knowledge of LLMs. 

\vskip 4pt
\noindent
\textbf{Reasoning.} The reasoning score for 0-shot CoT is marginally higher than for 2-shot CoT for both the original and CR-LT queries. We observe that with 2-shot CoT, the LLM at times erroneously follows the reasoning strategies in the examples when a different reasoning strategy is required. Furthermore, the reasoning scores, unlike accuracy, answer rate, and FaCTScore, do not decrease drastically in the long-tail setting. This is expected since substituting the entities in the queries does not fundamentally change the reasoning steps. 

\subsection{Claim Verification}

\vskip 4pt
\noindent
\textbf{Overall.} Similar to the question answering subtask, we observe a drastic decrease in accuracy and answer rate with the long-tail CR-LT queries. For instance, with 0-shot CoT, the accuracy drops from 0.89 to 0.35, and the answer rate drops from 0.93 to 0.44. These results reinforce the observation that LLMs struggle massively in long-tail settings.

\vskip 4pt
\noindent
\textbf{Factuality.} The FaCTScore for LLMs on our long-tail queries also leaves much room for future improvement. 0-shot CoT achieves a marginally higher FaCTScore (0.59) than 2-shot CoT (0.58). Coupled with the low FaCTScores in the question answering subtask as well as the decrease in FaCTScore when moving to the long-tail setting, these results reveal the propensity for LLMs to hallucinate, and thus the need for grounding answers in KGs.

\vskip 4pt
\noindent
\textbf{Reasoning.} 0-shot and 2-shot CoT achieve similar reasoning scores in both original and long-tail settings. Overall, there is a slight decrease in reasoning score in the long-tail setting, but the decrease is not nearly as significant as the drop in accuracy, answer rate, and FaCTScore. These results closely parallel those observed in the question answering subtask. 

%% file: Tables/results.tex
% Please add the following required packages to your document preamble:
% \usepackage{multirow}
\begin{table*}[t!]
\centering
\begin{tabular}{lllcccccccclcc}
\hline
\multirow{2}{*}{\textbf{Task}} &
  \multirow{2}{*}{\textbf{Model}} &
  \textbf{} &
  \multicolumn{2}{c}{\textbf{Accuracy}} &
  \textbf{} &
  \multicolumn{2}{c}{\textbf{Answer Rate}} &
   &
  \multicolumn{2}{c}{\textbf{FaCTScore}} &
   &
  \multicolumn{2}{c}{\textbf{Reasoning}} \\ \cline{4-5} \cline{7-8} \cline{10-11} \cline{13-14} 
 &
   &
   &
  \textbf{Original} &
  \textbf{CR-LT} &
   &
  \textbf{Original} &
  \textbf{CR-LT} &
   &
  \textbf{Original} &
  \textbf{CR-LT} &
   &
  \textbf{Original} &
  \textbf{CR-LT} \\ \cline{1-11} \cline{13-14} 
\multirow{2}{*}{\textbf{\begin{tabular}[c]{@{}l@{}}Question \\ Answering\end{tabular}}} &
  0-shot CoT &
   &
  0.70 &
  0.32 &
   &
  0.89 &
  0.41 &
   &
  0.63 &
  0.54 &
   &
  0.90 &
  0.89 \\
 &
  2-shot CoT &
   &
  0.70 &
  0.43 &
   &
  0.92 &
  0.57 &
   &
  0.64 &
  0.52 &
   &
  0.92 &
  0.90 \\ \cline{1-11} \cline{13-14} 
\multirow{2}{*}{\textbf{\begin{tabular}[c]{@{}l@{}}Claim \\ Verification\end{tabular}}} &
  0-shot CoT &
   &
  0.89 &
  0.35 &
   &
  0.93 &
  0.44 &
   &
  0.76 &
  0.59 &
   &
  0.93 &
  0.91 \\
 &
  2-shot CoT &
   &
  0.92 &
  0.41 &
   &
  0.97 &
  0.49 &
   &
  0.78 &
  0.58 &
   &
  0.93 &
  0.92 \\ \hline
\end{tabular}
\caption{Baseline results for both subtasks (question answering and claim verification) on both the original and CR-LT-KGQA queries. FaCTSCore and Reasoning are human-evaluated metrics.}
\label{results}
\end{table*}

%% file: CONTENT/conclusion.tex
In this work, we propose CR-LT KGQA, a novel KGQA dataset with two subtasks, question answering and claim verification, whose queries require both commonsense reasoning and long-tail knowledge. CR-LT is the first KGQA dataset whose queries require going beyond factoid retrieval but also making commonsense inferences. The lack of such KGQA datasets also means that existing KGQA methods only support factoid queries and thus do not support the queries in CR-LT. Instead, we evaluate LLMs using CoT, and find a high rate of hallucination in the long-tail setting. Thus, CR-LT KGQA represents a novel, challenging dataset for both hallucination-prone LLMs and existing KGQA methods limited to factoid queries. CR-LT paves the way for future KGQA research to design methodologies capable of answering the challenging commonsense and long-tail queries in the new era of LLMs.